\documentclass[conference]{IEEEtran}
\IEEEoverridecommandlockouts
\usepackage{cite}
\usepackage{amsmath,amssymb,amsfonts}
\usepackage{algorithmic}
\usepackage{graphicx}
\usepackage{textcomp}
\usepackage{amsfonts}
\usepackage{subfig}
\usepackage{multirow}
\usepackage{hyperref}
\usepackage[english]{babel}

\def\BibTeX{{\rm B\kern-.05em{\sc i\kern-.025em b}\kern-.08em
    T\kern-.1667em\lower.7ex\hbox{E}\kern-.125emX}}
\begin{document}

\title{Wildfire Detection Via Transfer Learning: A Survey\\

}
\author{
\IEEEauthorblockN{ Ziliang Hong, Emadeldeen Hamdan, Yifei Zhao, Tianxiao Ye, Hongyi Pan, A. Enis Cetin}
\IEEEauthorblockA{\textit{Department of Electrical and Computer Engineering} \\
\textit{University of Illinois Chicago}\\
Chicago,  IL 60607, USA \\
}
}

\maketitle
\begin{small}

\begin{abstract}
This paper surveys different publically available neural network models used for detecting wildfires using regular visible-range cameras which are placed on hilltops or forest lookout towers. The neural network models are pre-trained on ImageNet-1K and fine-tuned on a custom wildfire dataset. The performance of these models is evaluated on a diverse set of wildfire images, and the survey provides useful information for those interested in using transfer learning for wildfire detection. Swin Transformer-tiny has the highest AUC value but ConvNext-tiny detects all the wildfire events and has the lowest false alarm rate in our dataset.
\end{abstract}


\section{Introduction}
Early detection of wildfires is crucial in minimizing the harm they cause to people and the economy. Researchers have developed different techniques, including real-time algorithms that use video-based surveillance systems and deep neural networks that can recognize fire and smoke images\cite{guillemant2001real,toreyin2005wavelet,toreyin2006computer,yuan2008fast,toreyin2009wildfire,gunay2010fire,habiboglu2011real,habibouglu2012covariance,gunay2012entropy,ccetin2013video,gunay2015real,aslan2019early,jindal2021real,toreyin2006computer,cetin2016methods}. Some of these methods enable a single camera to detect wildfire smoke in real-time from a distance\cite{pan2020additive}.

Neural network-based methods for wildfire detection eliminate the need for manual feature selection, but they require a lot of data and computing power. To mitigate the issue of insufficient data, synthetic data is used for training, and transfer learning techniques like Fast R-CNN\cite{girshick2015fast} and Yolo series algorithms\cite{redmon2016you} are utilized to enhance the model's performance and reduce the amount of data required\cite{zhang2018wildland,wu2017adaptive}. Transfer learning has been used in this survey, inspired by previous papers\cite {pan2020computationally,pan2020additive}, leading to the development of effective neural network models for forest fire detection since 2015.

In this paper, we explore the feasibility of several classical models in the field of wildfire detection and compare their performance. These models are Residual Neural Network V2(ResNetV2), Data-efficient, image Transformers (DeiT), EffecientNetV2, Big Transfer (BiT), MobileNetV3, Swin Transformer, and ConvNeXt\cite{he2016identity,hinton2015distilling,touvron2021training,tan2021efficientnetv2,kolesnikov2020big,howard2019searching,liu2021swin,liu2022convnet}. Inspired by a previous paper \cite{pan2020additive}, a single image is split into sub-images for object detection. We evaluate critical indicators, including accuracy, false alarm rate, true detection rate, detection latency, and implementation latency for wildfire detection \cite{pan2020additive,pan2020computationally,pan2022deep}. Experiments compare model performance, analyze superior models, and provide a summary of comparisons.

\section{Overview of Artificial Neural Networks}
In this section, we present a comprehensive review of the selected models, detailing their respective structures and innovations.

\subsection{Residual Neural Network (ResNet)}

ResNet is a revolutionary neural network \cite{he2016deep}. 
It achieved unprecedented depth with more than 1000 layers, owing to the introduction of residual blocks \cite{he2016deep}. Before the ResNet, it is always challenging to design very deep Convolutional Neural Networks (CNNs) because neural networks that are too deep usually face the problems of gradient vanishing and exploding. Besides, an increasing number of layers could also cause degeneration. Through the skip connection structure of layers which is called residual block and the usage of the Batch Normalization (BN) layer, ResNet shows its ability to solve this predicament. The core equation of a residual block in ResNet can be expressed as follows:
\begin{flalign}
   &&
  y=\mathcal{F}(x, \{W_i\})+W_s x.
   &&
\end{flalign}
  
In ResNetv2 \cite{he2016identity}, the structure of the residual module was improved. The BN layer and activation function are placed in front of the weight layer as preactivation. This is the main difference between ResNetv2 and ResNetV1. Such a structure can not only afford efficient backpropagation but also allow the BN layer to play a regularization role, which makes ResNetV2 significant progress.

\subsection{MobileNet}
MobileNet \cite{howard2017mobilenets} is built primarily from depthwise separable convolutions. Depthwise separable convolutions include a standard convolution into a depthwise convolution and a 1×1 convolution called a pointwise convolution. It can reduce computation and model size effectively. 



Meanwhile, the network structure has been optimized and a parameter called width multiplier has been introduced to further thin the network uniformly at each layer. MobileNet is small and has low latency. It is implementable on computationally limited platforms and shows strong performance. The authors introduce MobileNetV2 using the inverted residual with the linear bottleneck to further improve the network \cite{sandler2018mobilenetv2}.  
In the next generation MobileNetV3, a combination of these modules (depthwise separable convolutions and inverted residual with linear bottleneck) are used in the first two generations. Platform-aware Neural Architecture Search (NAS) is also employed to search for the global network structures and then use the NetAdapt algorithm to search per layer for the number of filters. Computationally expensive layers has been redesigned and a nonlinearity called swish to replace ReLU is used instead. As the next generation of MobileNet, MobileNetV3 can achieve higher accuracy and lower latency than MobileNetV2 \cite{howard2019searching}. 

\subsection{Big Transfer (BiT)}

The innovation of BiT lies in its large-scale pretraining strategy that involves training on multiple public datasets. By leveraging this strategy, BiT achieves high performance on a wide range of computer vision tasks and surpasses previous state-of-the-art results. Also, Group Normalization and weight standardization is used instead of Batch Normalization. Because it incurs inter-device synchronization costs when using distributed training. And it is detrimental to transfer due to the requirement to update running statistics.

During the transfer to downstream tasks, a fine-tuning protocol called BiT-HyperRule is proposed. It is heuristic to set the following hyperparameters per-task: training schedule length, resolution, and whether to use MixUp regularization. The models are evaluated on standard benchmarks and have a good performance. The recipe is simple and effective when we transfer pre-trained models to diverse tasks \cite{kolesnikov2020big}.

\subsection{EffecientNet}
 The compound scaling method is proposed to scale network width, depth, and resolution with a set of fixed scaling coefficients. NAS is used to design a new baseline network and scale it up to obtain a family of models called EfficientNets, which is smaller and faster than existing convolutional neural networks \cite{tan2019efficientnet}.  

EfficientNetV2 is introduced in June 2021. A combination of training-aware NAS and scaling are used to improve both training speed and parameter efficiency. They design a search space enriched with additional ops such as Fused-MBConv and propose an improved method of progressive learning, which can adjust regularization along with image size. EfficientNetV2 have up to 11x faster training speed and up to 6.8x better parameter efficiency on ImageNet, CIFAR, Cars, and Flowers dataset, than prior art such as ResNet-101 and ViT-L/16 (21k) \cite{tan2021efficientnetv2}.

\subsection{Data-efficient image Transformers (DeiT)}
Recurrent Neural Networks (RNNs) require information from previous or next-time steps for calculations, making parallel computation difficult and limiting them to serial processing. By adopting the Self-Attention mechanism \cite{vaswani2017attention}, The Transformer model for sequence processing avoids horizontal propagation, relying instead on vertically stacked self-attention layers that allow for parallel computation and acceleration using GPUs.

Vision Transformer (Vit) is a transformer-based model in computer vision which require massive training data \cite{dosovitskiy2020image}. In order to overcome the limitations of Vit, the Data-efficient image transformers (DeiT) model was developed. A systematic optimization and regularization approach was employed on DeiT, which includes data augmentation during training. The authors employed the so-called soft and hard-label knowledge distillation to facilitate the teacher model in guiding DeiT during training \cite{hinton2015distilling,touvron2021training}.

 %
\subsection{Swin Transformer}

Swin Transformer \cite{liu2021swin} uses a hierarchical construction method similar to CNNs. Such a backbone helps to build detection and segmentation tasks on this basis. When using the Windows Multi-head Self-Attention (W-MSA) module, the self-attention calculation will only be performed within each window, so there is no information transfer from window to window. To solve this problem, the authors introduced the Shifted Windows Multi-Head Self-Attention (SW-MSA) module. Relative Position Bias is also employed to improve the performance of the model.

\subsection{ConvNeXt}
ConvNeXt does not introduce novel architectural or methodological innovations. Instead, it leverages existing techniques and optimized CNNs for enhanced performance. The authors first used the strategy of training ViT to train the original ResNetV2-50 model and observed a significant improvement in performance compared to the baseline. This benchmark performance was then utilized for subsequent experiments. Through a series of experiments, ConvNeXt has faster inference speed and higher accuracy compared with Swin Transformer with the same computational complexity \cite{liu2022convnet}.
\section{Methodology}

\subsection{Methods of Implementation}

Object detection tasks involve identifying the location and classification of objects in images. Usually, the bounding box-based method needs to manually label massive bounding boxes like the Yolo series algorithm\cite{redmon2016you}. In this experiment, the approach taken is to divide each image into 45 blocks as Fig. \ref{Methods Exmaple}. Using this approach, it is possible to detect and locate fires accurately and we do not need to label the bounding box of the wildfire. Assuming that the dimensions of the active image area are $M_i$ and $N_i$ and that the dimensions of each block are $M_b$ and $N_b$, an equation can be used to express the row number R and column number C for each block:
\begin{equation}
 R=\lfloor \frac{M_i}{M_b}\rfloor, C=\lfloor \frac{N_i}{N_b}\rfloor   
\end{equation}
where $\lfloor\cdot\rfloor$ stands for the floor function.
The main task is to construct a binary classifier that can predict the whether there is a wildfire or not. Forest fire detection devices are often used in remote wilderness areas where weight and computational resources are limited. Therefore, small models are chosen to fit these devices.

\begin{figure}[htbp]
\centering
\includegraphics[width=1\linewidth]{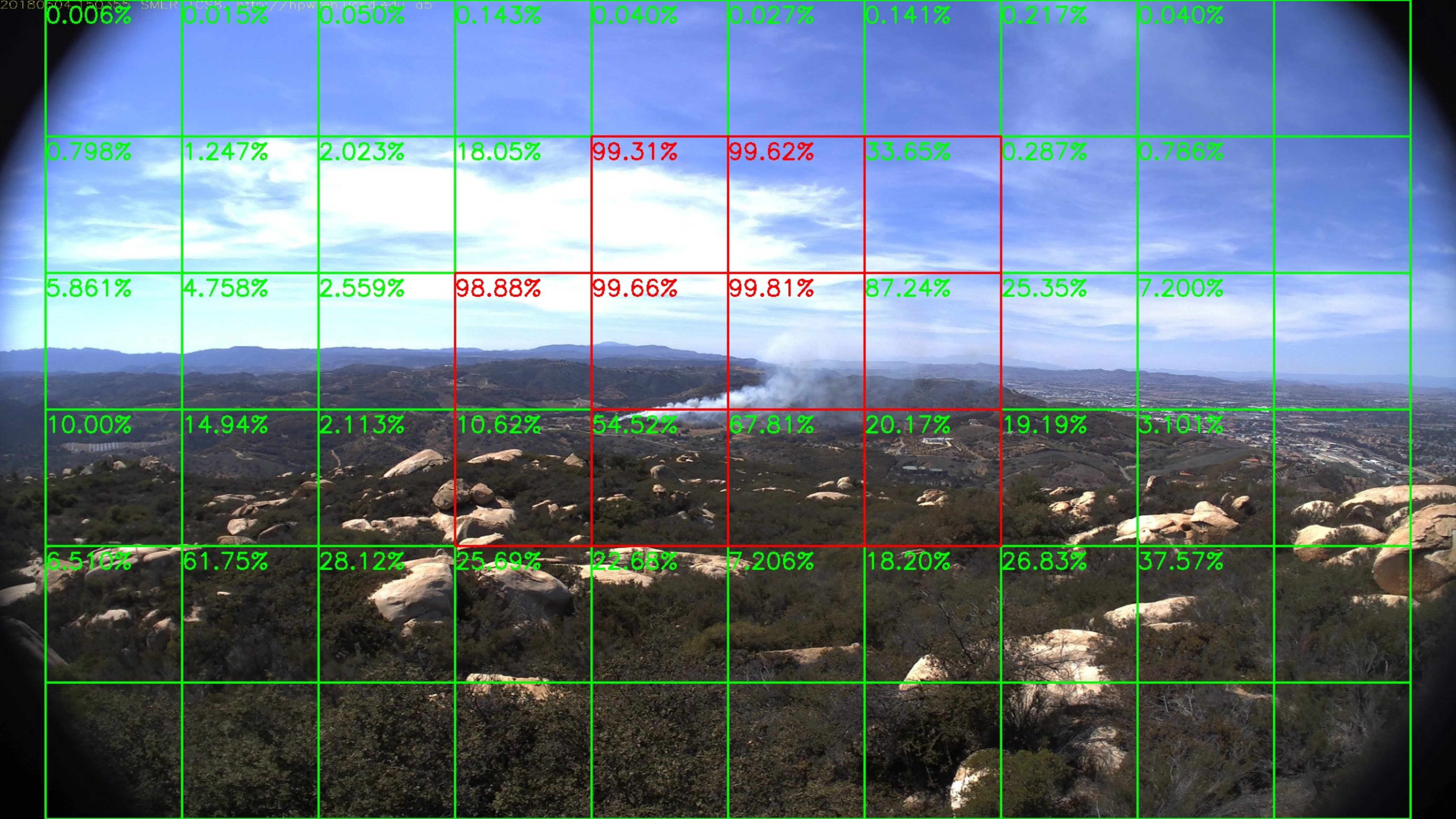}
\caption{The top is a sample to be detected, and the below is the detected result. (detected by Swin Transformer-tiny)}
\label{Methods Exmaple}
\end{figure}

\subsection{Dataset}
In this work, we build a wildfire dataset by enriching the dataset in~\cite{pan2020additive} and~\cite{pan2020computationally} to about 35k images. The training subset approximately contains 14k images of normal forests and 9k wildfire images. The test subset contains about 8k images of normal forests and 4k wildfire images. All the data images are from the HPWREN wildfire dataset \cite{UCSD}, the FIRESENSE database~\cite{grammalidis2017firesense}, google images, and YouTube videos. Furthermore, to evaluate the performance in real-world applications, we evaluate the models on the HPWREN videos to estimate the detection latency. Fig. \ref{Dataset Example} and Fig. \ref{Exmple Of Detection Delay} show some samples from the dataset.

\begin{figure}[htbp]
\centering
\subfloat[]{\includegraphics[height=2cm,width=2cm]{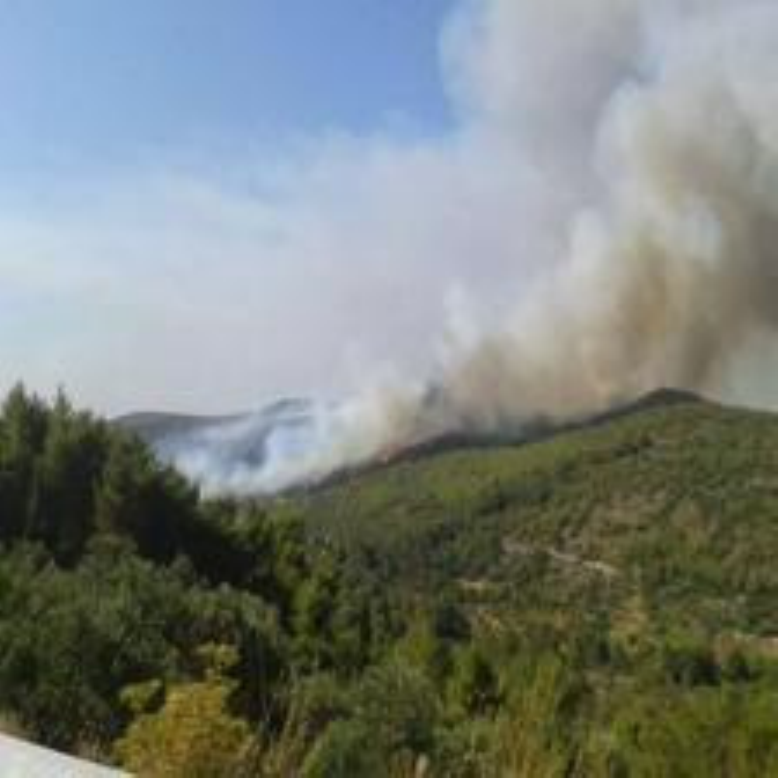}}
\subfloat[]{\includegraphics[height=2cm,width=2cm]{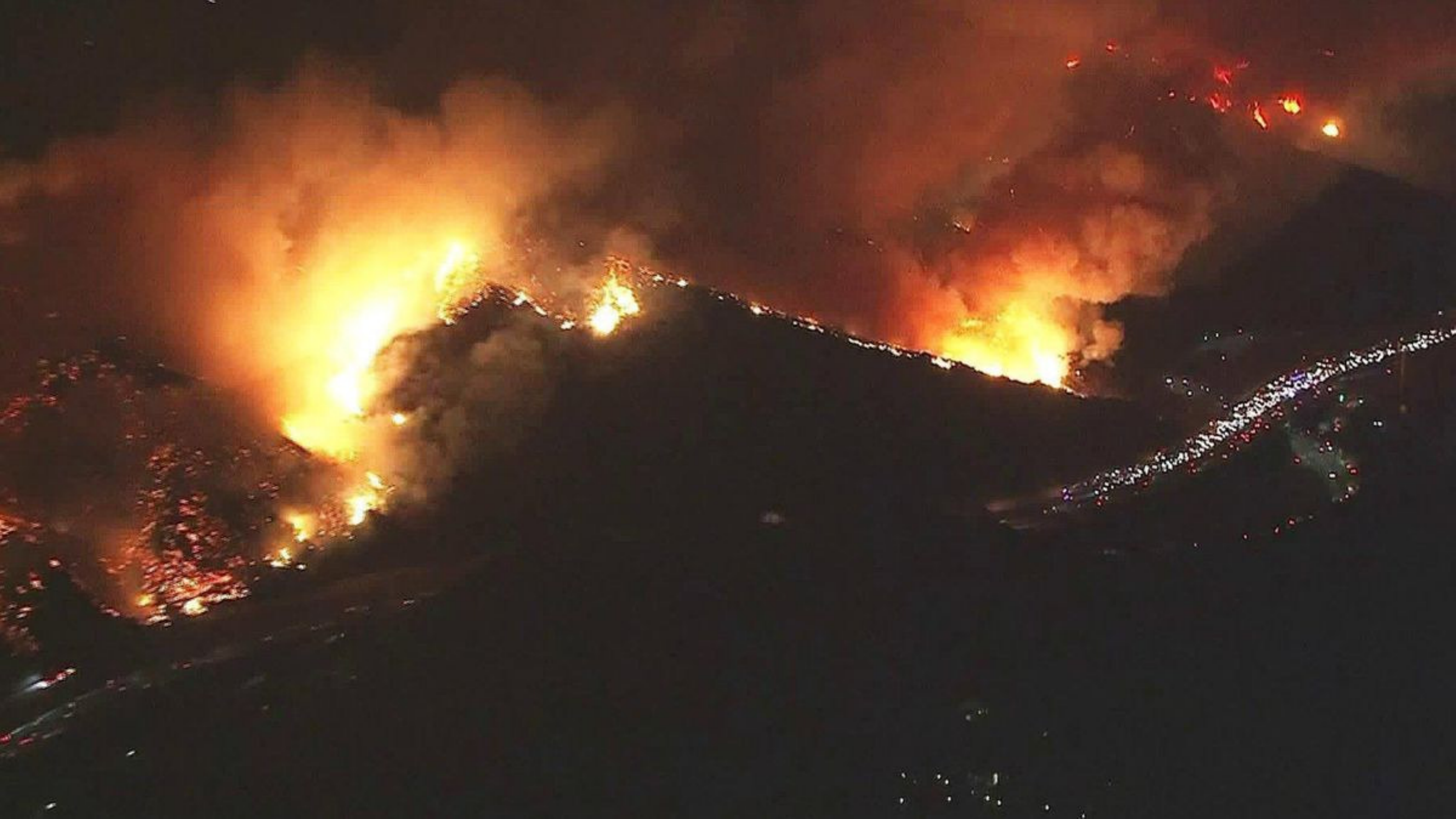}}
\subfloat[]{\includegraphics[height=2cm,width=2cm]{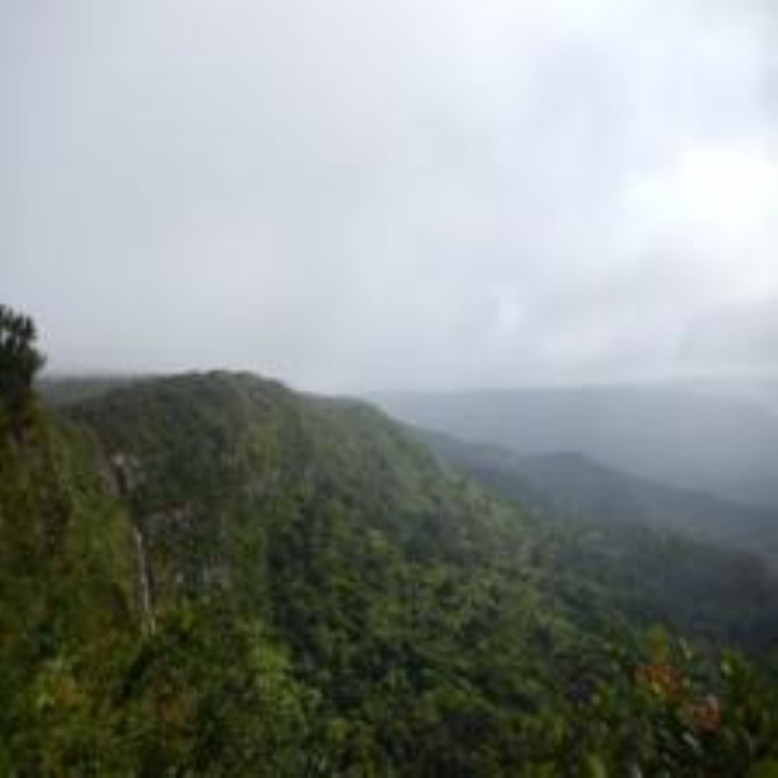}}
\subfloat[]{\includegraphics[height=2cm,width=2cm]{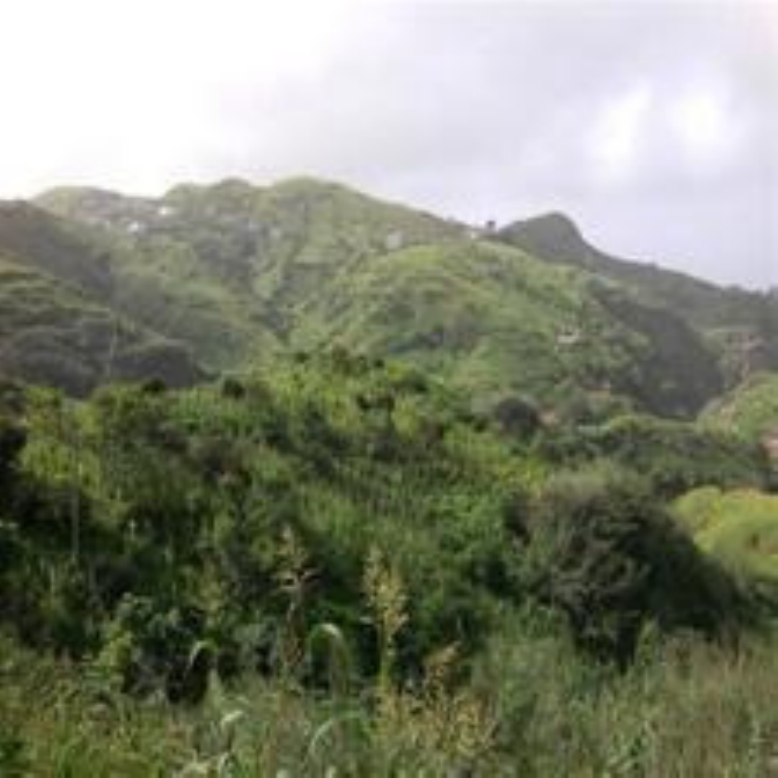}}
\caption{Examples of the dataset images. Wildfire exists in (a) and (b). Wildfire does not exist in (c) and (d).}
\label{Dataset Example}
\end{figure}

\begin{figure}[htbp]
\includegraphics[width=0.48\linewidth]{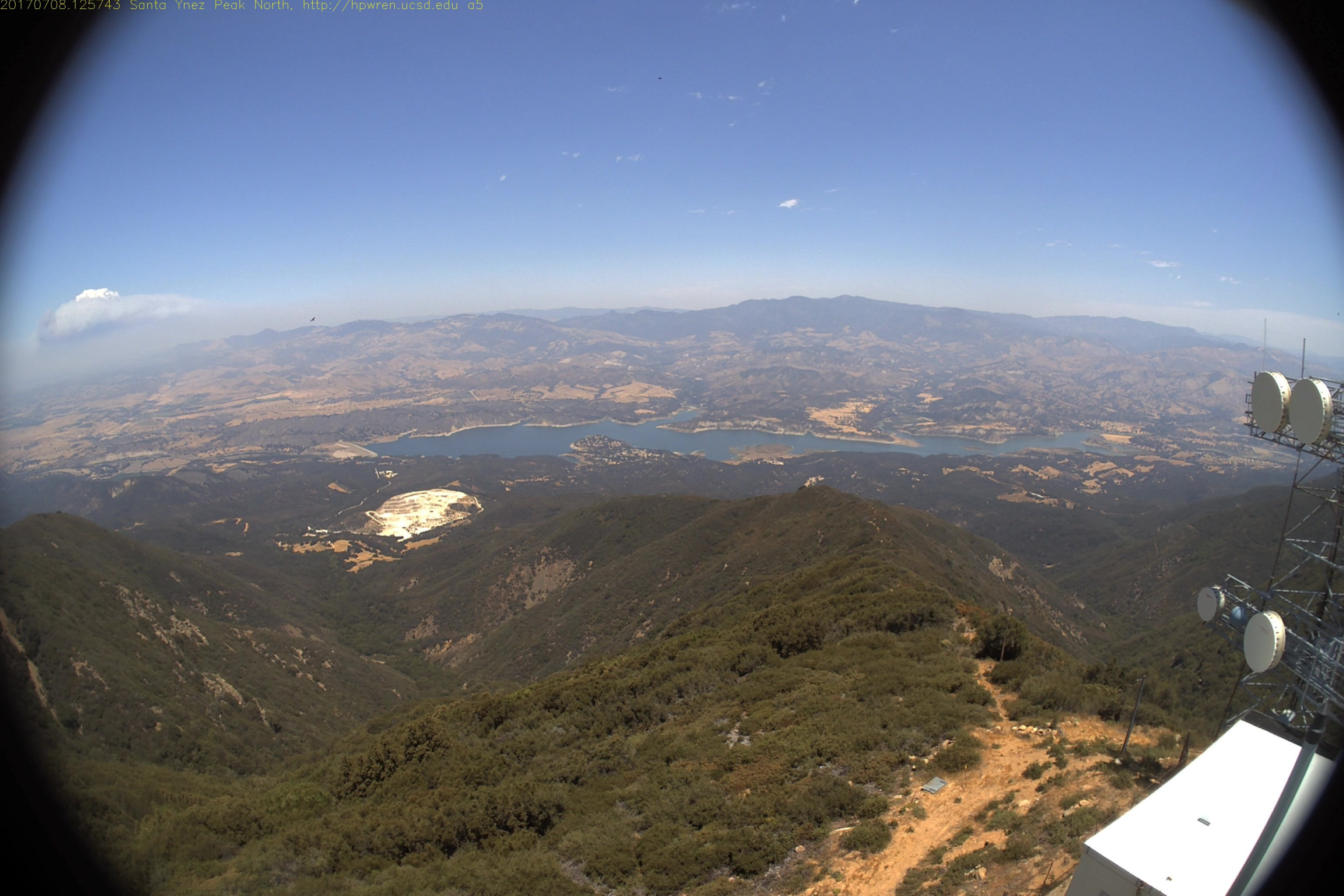}
\includegraphics[width=0.48\linewidth]{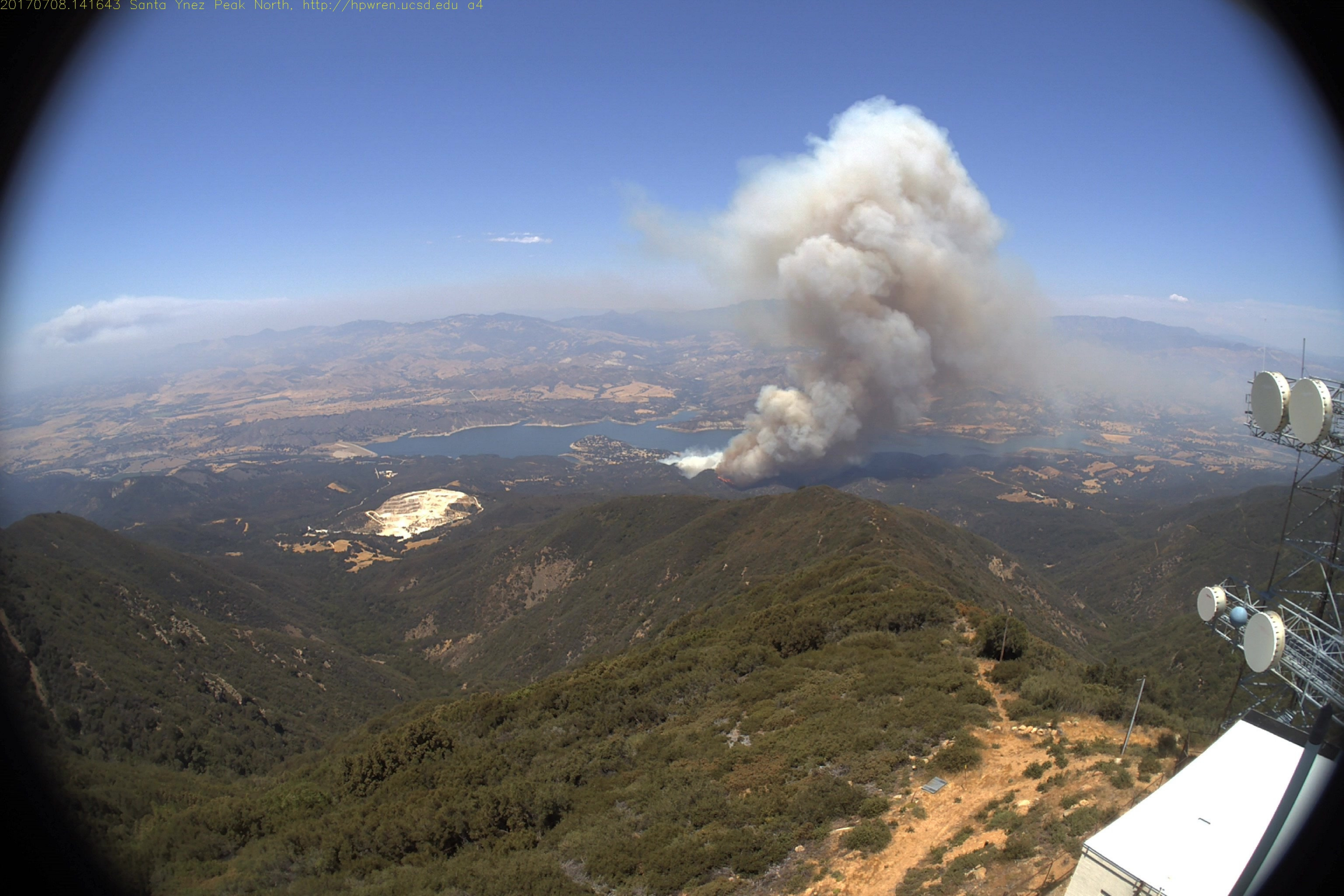}

\includegraphics[width=0.48\linewidth]{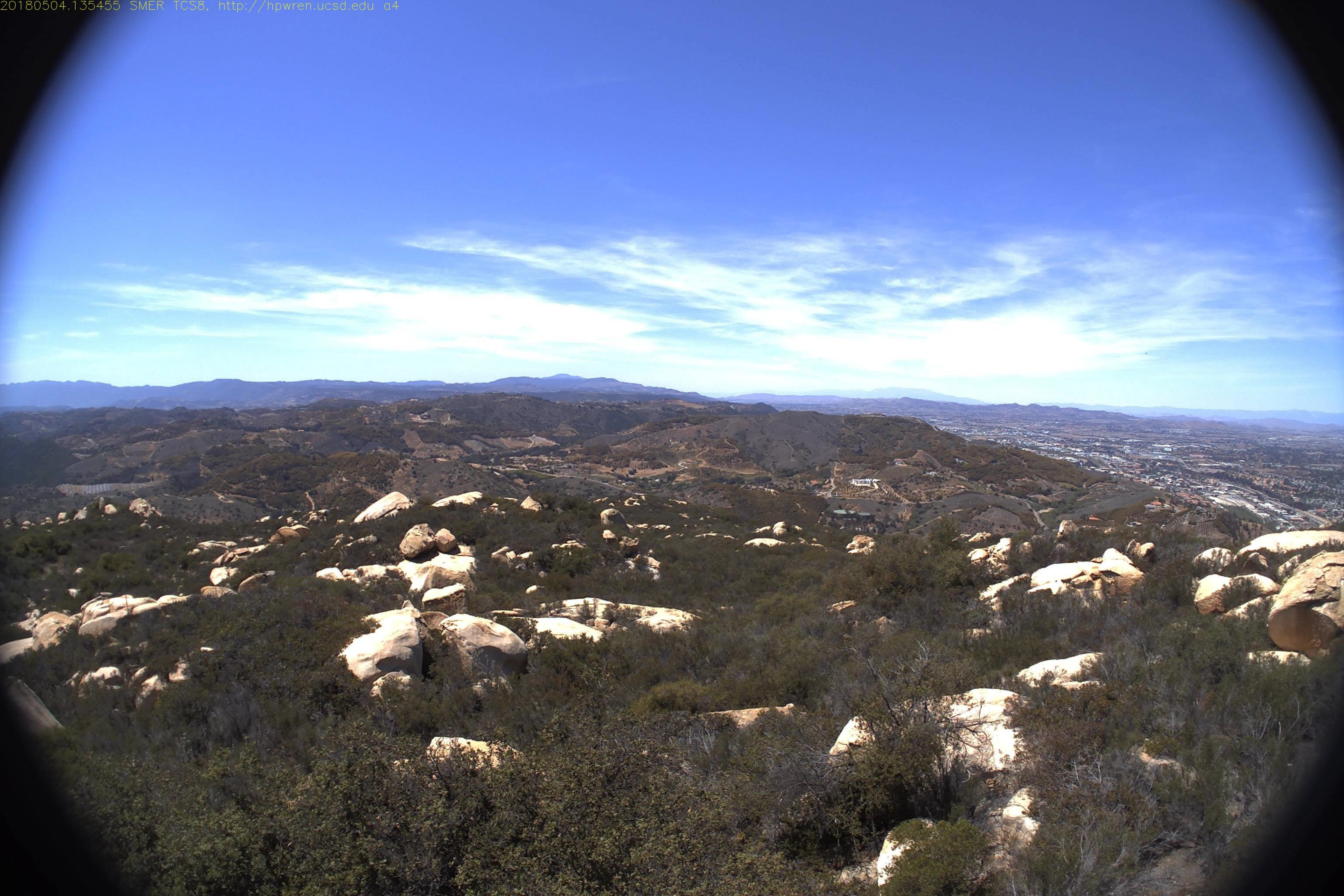}
\includegraphics[width=0.48\linewidth]{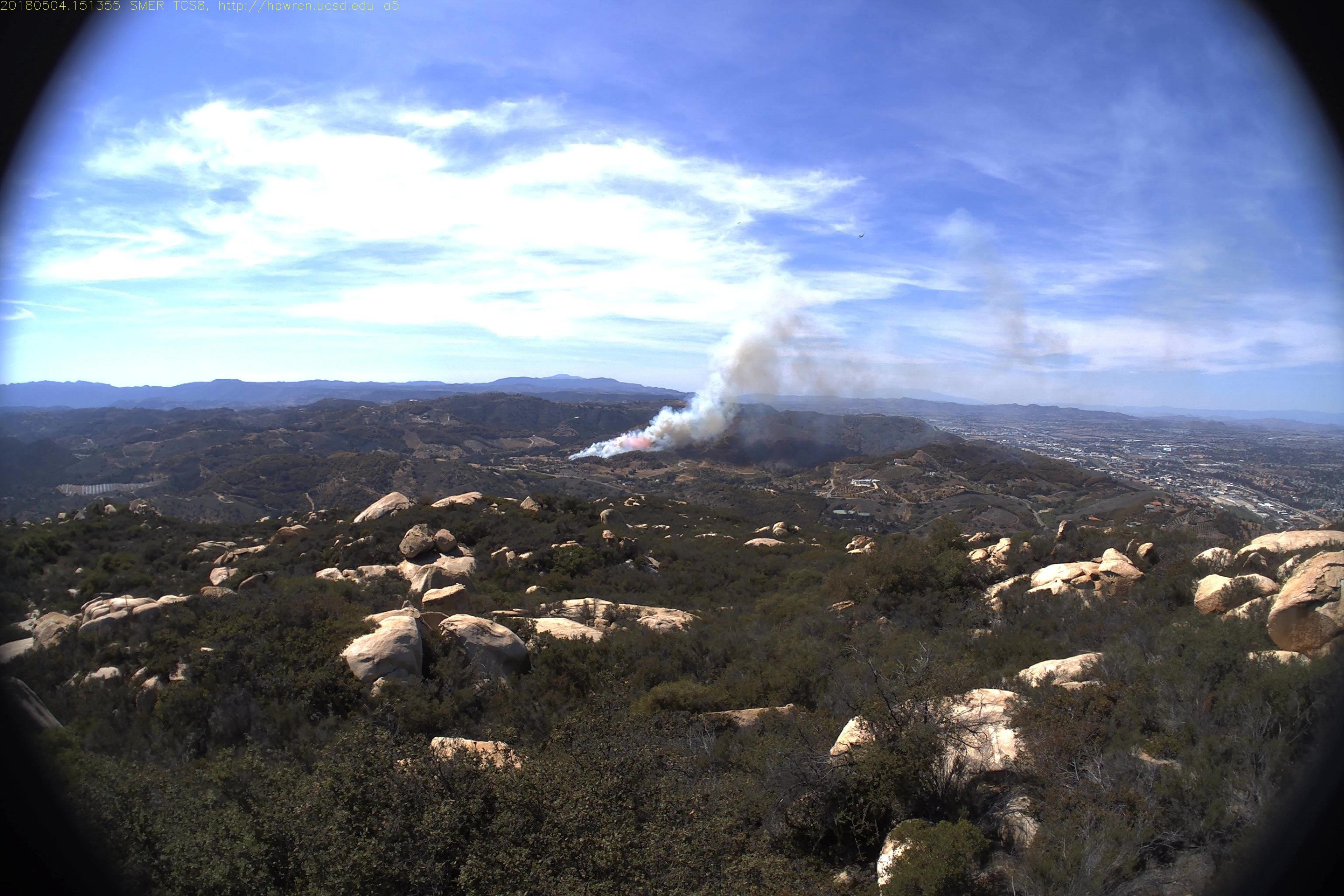}
\caption{HPWREN samples for test model detection. We use data from 9 HPWREN cameras, each of which recorded the process of occurrence of fire.}
\label{Exmple Of Detection Delay}
\end{figure}

\section{Experiment}

\subsection{Training Models}
Transfer learning techniques in deep learning can address the time-consuming task of gathering extensive training data and overfitting problems 
In this study, pre-trained models from the ImageNet-1K open-source database~\cite{deng2009imagenet} were used for forest fire detection using theTensorFlow deep learning library with an NVIDIA RTX3070 GPU. The training process consisted of 15 epochs with a fixed feature extractor method used in the first 10 epochs and a transfer learning strategy applied in the final 5 epochs. Fine-tuning all the layers in the whole model further improved the models' performance while reducing the demand for large amounts of data. After fine-tuning, the probability of wildfire was calculated by feeding the results to a softmax layer. The study assumed that images with fire are negative samples and images without fire are positive samples. To ensure practical application, the wildfire detection model should not be overly sensitive while accurately detecting forest fires to the greatest extent possible using the true detection rate and false alarm rate in the confusion matrix.

\begin{figure}[htbp]
\centering
\includegraphics[width=\linewidth]{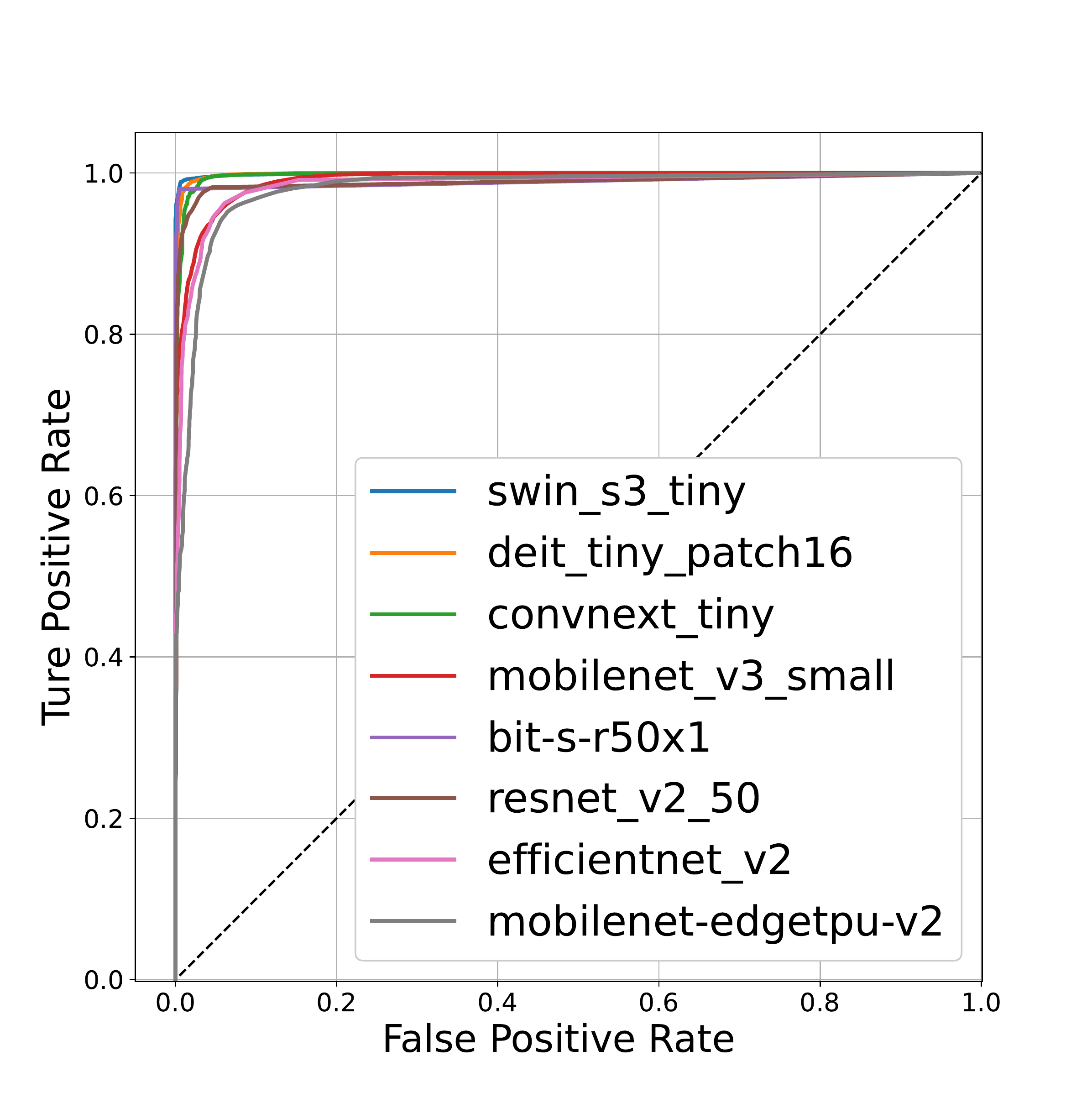}
\caption{ROC of models. Models are arranged by AUC size from top to bottom. }
\label{fig: ROC}
\end{figure}
\begin{table*}[htbp]		
\centering
		\begin{tabular}{lcccccc}
			\hline\noalign{\smallskip}
   \textbf{Model}&\textbf{Parameters (M)}&\textbf{FLOPs (G)}&\textbf{AUC}&\textbf{ACC}&\textbf{True Detection}&\textbf{False Alarm}\\
			\noalign{\smallskip}\hline\noalign{\smallskip}
			Swin Transformer-tiny  & 27.6 &8.99 & \textbf{0.99917}   & 97.95$\%$   & 94.63$\%$  & 0.36$\%$   \\
                DeiT-tiny  &5.5 & 2.54 & 0.99876& \textbf{98.13}$\%$  &95.12$\%$  & 0.35$\%$  \\
                ConvNeXt-tiny  & 27.8  & 8.99 & 0.99743 
             & 96.69$\%$   & 90.52$\%$  & 0.18$\%$\\
                MobileNetV3-small & 20.3   & \textbf{0.12} & 0.99114   & 95.17$\%$   & 94.11$\%$  & 4.30$\%$\\
BiT-small & 23.5   & 8.38 & 0.98973  & 96.86$\%$   & \textbf{99.78}$\%$ & 4.63$\%$   \\
ResNetV2-50 & 23.6   & 6.99 & 0.98837   & 95.00$\%$   & 98.92$\%$  & 6.99$\%$ \\
EfficientNetV2 & \textbf{1.5}   & 5.74 & 0.98571 & 93.47$\%$   & 96.48$\%$  & \textbf{8.06}$\%$    \\
Mobilenet-edgetpu-v2 & 2.5   & 1.03 & 0.98056  & 94.70$\%$   & 91.73$\%$  & 3.79$\%$  \\
   
			\noalign{\smallskip}\hline
		\end{tabular}
            \caption{Parameters, FLOPs, AUC of models. Accuracy, True Detection Rate, and False Alarm Rate
of models on the test dataset. Models are arranged in descending order of AUC. }
		\label{FLOPs}
	\end{table*}

 \begin{table*}[htbp]
\centering 
\begin{tabular}{cccccccccccc}
	\hline\noalign{\smallskip}
\multicolumn{2}{l}{\textbf{Video No.}}
& \textbf{1}         
& \textbf{2}  
& \textbf{3}  
& \textbf{4}  
& \textbf{5}  
& \textbf{6}  
& \textbf{7}  
& \textbf{8}  
& \textbf{9} 
& \textbf{Average}\\ 
	\noalign{\smallskip}\hline\noalign{\smallskip}
\multicolumn{2}{l}{Swin Transformer-tiny}&  7          &  4  & \textbf{0} &   27 &  6 &  \textbf{1}  &  2 &  4  & 2 & 5.8\\ 

\multicolumn{2}{l}{DeiT-tiny}& 5           &      4     &  1 &  29  & 7  & 2  &  6 &  4  & 2 & 6.7\\ 

\multicolumn{2}{l}{ConvNeXt-tiny}&  5          &      4     &  \textbf{0} & 29   & 7  &  \textbf{1}  &  6 & 6   & \textbf{1} & 6.6\\

\multicolumn{2}{l}{MobileNetV3-small} & 6  &      11     &  \textbf{0} &  35  &  7 &  4  &  27 &   13 & \textbf{1} & 11.6\\ 

\multicolumn{2}{l}{BiT-small}&     \textbf{2}      &      \textbf{0}     & \textbf{0}  &  \textbf{26}  &  \textbf{4} & \textbf{1}   &  \textbf{1} &  3  & \textbf{1} & \textbf{4.2}\\ 

\multicolumn{2}{l}{ResNetV2-50}&   3        &       2    &  \textbf{0} &  29  &  5 & \textbf{1}   &  8 & 7   & 2 & 6.3\\ 

\multicolumn{2}{l}{EfficientNetV2}& 4   &  2 &  \textbf{0} &  26  & 6  & 4  & 6  &  \textbf{2}  & \textbf{1} & 5.7\\ 
\multicolumn{2}{l}{Mobilenet-edgetpu-v2}& 6  &     11      & 1  &   35 &  7 &  4  & 6  & 11   & \textbf{1} &9.1 \\ 
	\noalign{\smallskip}\hline
\end{tabular}
\caption{Considering the rigor of fire prediction, we set the threshold to 95$\%$. Each number represents the frame number at which the models first detected the presence of fire.We Calculate detection latency as the time between the frame when the fire is first detected and the frame when the fire starts. In the last column, we calculate the average detection latency.}
\label{detection delay}
\end{table*}
\subsection{Testing Models}
\subsubsection{Evaluation Indicators}
The evaluation of deep neural networks (DNNs) involves five key indicators: accuracy, true detection rate, false alarm rate, floating point operations (FLOPs), number of parameters, detection latency, and Receiver Operating Characteristic (ROC). ROC curve is a graph that shows how sensitive a model is to various threshold ranges between 0$\%$ and 100$\%$ \cite{woods1997generating}. By calculating the False Positive Rate (FPR) and True Positive Rate (TPR) under different thresholds and setting them as X-axis and Y-axis respectively, ROC curve and Area Under the Curve (AUC) can be used to compare the performance of different models and usually a larger AUC indicates a better performance. The mathematical expressions for ROC and AUC can be represented as:

\begin{equation}
\text{ROC} = \frac{\text{TPR}}{\text{TPR} + \text{FPR}}
\end{equation}

\begin{equation}
\text{AUC} = \int_0^1 \text{ROC}(x) \, \mathrm{d}x
\end{equation}

Accuracy measures the percentage of accurate predictions generated by the neural network, indicating the models' fitness for purpose. True detection rate and false alarm rate are crucial indicators in wildfire detection systems, with true detection rate representing the proportion of correct negative predictions made by the model and false alarm rate indicating the percentage of false alarms, with a lower false alarm rate indicating greater reliability.

Parameters and FLOPs are essential factors affecting neural networks, with the former indicating the model's complexity and the latter measuring the number of calculations an algorithm can perform in a second. FLOPs are also used to estimate the computational requirements of training or executing a neural network on a hardware device.

Finally, detection latency refers to the interval between the start of an actual fire and the system's detection of it. In experiments, this metric is evaluated in frames rather than seconds. These indicators can assist in evaluating the potential applicability and performance of DNNs, which can inform the development of more reliable and efficient models.

%

\subsubsection{Basic Test}

Fig. \ref{fig: ROC} shows the ROC curve of each model. Because the curves are dense, Fig. \ref{fig: ROC} displays the details of the top left part.
Table \ref{FLOPs} shows the FLOPs and parameters of each model, the models are arranged in the order of AUC size, from largest to smallest. Table \ref{FLOPs} records the performance of each model on the test dataset with the threshold of 95$\%$. As Table \ref{FLOPs} shows, DeiT-tiny owns the highest accuracy which is 98.13$\%$ followed by 97.95$\%$ from Swin Transformer-tiny and they are both transformer-based models. The remaining traditional CNNs models perform slightly worse, with BiT-small and ConvNeXt-tiny performing better than 96$\%$, the rest of the models' performance is relatively mediocre. 

For image recognition with fire, traditional CNNs perform very well, with ResNetV2-50 reaching 98$\%$ accuracy and BiT-small even reaching 99$\%$. However, a high true detection rate comes at the cost of a high false alarm rate. Both models have a high false alarm rate, ResNetV2-50 was found to have a false alarm rate of 6.99$\%$. On the contrary, the transformer-based models have a very low false alarm rate, although the accuracy is slightly lower than that of the CNNs models. ConvNeXt-tiny, a member of the traditional CNNs, has a slightly lower accuracy than other CNN models but has a comparable false alarm rate with the transformer-based models.

\subsubsection{Detection Latency}

 Detection latency is a way to describe the time from the onset of the fire until the model raises an alarm and it is measured in frames. Nine datasets are selected to calculate the detection latency, which does not overlap with the data in the test dataset. The result is shown in Table \ref{detection delay}.

In Table \ref{detection delay}, BiT-small demonstrates the best performance, exhibiting a remarkable ability to recognize small features. However, this also leads to a higher susceptibility to similar features, which is related to  BiT-small's higher false alarm rate in Table \ref{FLOPs}. The performance of ConvNeXt-tiny and DeiT-tiny in the detection latency test is moderate, despite their good inference speed and accuracy. Moreover, the difference in detection latency between them is insignificant. On the other hand, Swin Transformer-tiny is slower in detection latency.

\subsubsection{Implement Latency}
\begin{table}[htbp]
\centering
\begin{tabular}{lcc}
\hline\noalign{\smallskip}
\textbf{Model}                       & \textbf{Sub-image} & \textbf{Whole Image}  \\
	\noalign{\smallskip}\hline\noalign{\smallskip}
Swin Transformer-tiny       & 0.02227              & 1.0023                         \\
DeiT-tiny
         & 0.02272              & 1.0223                         \\
ConvNeXt-tiny
      & 0.06151              &2.7680                         \\
MobileNetV3-small & 0.00591             & 0.2660                         \\
BiT-small                      & 0.01636                       & 0.7363                                  \\
ResNetV2-50                 & \textbf{0.00683}                       & \textbf{0.3074}                                  \\
EfficientNetV2               & 0.01441                       & 0.6483                                  \\
Mobilenet-edgetpu-v2        & 0.00892                       & 0.4013                                  \\

	\noalign{\smallskip}\hline
\end{tabular}
\caption{The time it takes for the model to process the image to make a prediction. Measured in seconds.}
\label{table3}
\end{table}
In addition to the model's ability to detect fires at the earliest stages,  we are also interested in image processing speed. Implement latency is used to measure the time that a model needed to process a single image. Since we split our single image into 45 sub-images, we both calculate the time of processing a single image and a sub-image. At the same time, Table \ref{table3} indicates the speed of models and we can note that MobileNetV3-small has the shortest implement latency and ResNetV2-50 has the second shortest implement latency. While ConvNeXt-tiny has a smaller false alarm rate compared to Swin-tiny and DeiT-tiny, it is not as efficient in processing images as these transformer-based models.

\section{Conclusion}
In this paper, we chose 8  pre-trained models that are widely used in image classification to validate the effectiveness and performance of artificial deep neural networks for wildfire detection using regular visible-range cameras. Transfer learning is used to train these models and the models are evaluated based on different indicators. Based on the experiments summarized above, the strengths and limitations of these models are determined. According to the criteria of traditional image classifiers, transformer-based models  Swin Transformer and DeiT achieved the highest AUC and accuracy. Traditional CNNs models have slightly lower AUC and accuracy. On the other hand, traditional CNN models except ConvNeXt are more efficient in implementation and superior in detecting tiny smoke features. Swin Transformer and DeiT consume more time to implement relatively, but it is still acceptable in practice because the processing time of an entire image frame size is less than 1 second except for Swin Transformer and DeiT networks.
Unfortunately, Swin Transformers and Deit are less sensitive to tiny smoke features. ConvNext-tiny has the lowest false alarm rate. A low false alarm rate is very important for the acceptance of the use of machine learning in wildfire detection problems.

Future research will focus on the use of other types of data, such as thermal and multispectral imagery. Additionally, the development of hybrid models that combine deep learning algorithms with traditional machine learning techniques and domain knowledge can also be a promising direction for improving the accuracy of wildfire detection systems. 
\end{small}
\bibliographystyle{unsrt}
\bibliography{reference}
\end{document}